\pdfoutput=1

\documentclass[11pt]{article}

\usepackage[final]{acl}

\usepackage{times}
\usepackage{latexsym}
\usepackage{amsmath}
\usepackage{multirow}
\usepackage{booktabs}
\usepackage{amssymb}
\usepackage{bbding}
\usepackage{pifont}
\usepackage{utfsym}
\usepackage{footmisc}

\usepackage{xcolor}
\definecolor{mygray}{rgb}{0.9,0.9,0.9}

\usepackage[T1]{fontenc}

\usepackage[utf8]{inputenc}

\usepackage{microtype}

\usepackage{inconsolata}

\usepackage{graphicx}

\newcommand*\samethanks[1][\value{footnote}]{\footnotemark[#1]}

\title{RAVEN++: Pinpointing Fine-Grained Violations in Advertisement Videos with Active Reinforcement Reasoning}

\author{
Deyi Ji$^1$\thanks{The first two authors contribute equally to this work.} \hspace{0.5em} Yuekui Yang$^{1,2}$\samethanks \hspace{0.5em} Liqun Liu$^1$\thanks{Corresponding Authors.}   \hspace{0.5em} Peng Shu$^1$   \hspace{0.5em} Haiyang Wu$^1$  \hspace{0.5em} Shaogang Tang$^1$ \\  \hspace{0.5em} \textbf{Xudong Chen}$^1$  \hspace{0.5em} \textbf{Shaoping Ma}$^2$   \hspace{0.5em} \textbf{Tianrun Chen}$^3$ \hspace{0.5em} \textbf{Lanyun Zhu}$^4$\samethanks \\
  $^1$Tencent \hspace{0.5em} $^2$Department of Computer Science and
Technology, Tsinghua University   \\ $^3$Zhejiang University  \hspace{0.5em} 
$^4$Nanyang Technological University  \\
\texttt{\{deyiji,yuekuiyang,liqunliu,archershu,gavinwu,teetang,seadenchen\}@tencent.com,} \\
\texttt{msp@tsinghua.edu.cn,tianrun.chen@zju.edu.cn,lanyun.zhu@ntu.edu.sg} 
}

\begin{document}
\maketitle

\begin{abstract}
Advertising (Ad) is a cornerstone of the digital economy, yet the moderation of video advertisements remains a significant challenge due to their complexity and the need for precise violation localization. While recent advancements, such as the RAVEN model, have improved coarse-grained violation detection, critical gaps persist in fine-grained understanding, explainability, and generalization. To address these limitations, we propose RAVEN++, a novel framework that introduces three key innovations: 1) Active Reinforcement Learning (RL), which dynamically adapts training to samples of varying difficulty; 2) Fine-Grained Violation Understanding, achieved through hierarchical reward functions and reasoning distillation; and 3) Progressive Multi-Stage Training, which systematically combines knowledge injection, curriculum-based passive RL, and active RL. Extensive experiments on both public and proprietary datasets, on both offline scenarios and online deployed A/B Testing, demonstrate that RAVEN++ outperforms general-purpose LLMs and specialized models like RAVEN in terms of fine-grained violation understanding, reasoning capabilities, and generalization ability.

\end{abstract}

\section{Introduction}

Advertising (Ad) plays a pivotal role in the digital economy, serving as a primary driver of revenue and growth for online platforms \cite{rathee2024sustainability,campbell2025diversity,ji2025hierarchical}. To maintain legal compliance, support sustainable development, and create a positive user experience, platforms implement strict content moderation policies for advertisers. Despite these efforts, violations of Ad guidelines remain prevalent, presenting ongoing challenges for effective moderation. While recent advancements in large language models (LLMs) \cite{llava,qwen,qwen_vl,zhu2024ibd,zhu2024llafs,zhu2025llafs++} have improved the detection of non-compliant content, critical gaps persist, especially in the moderation \cite{qiao2024scaling,madio2025content, hasan2024religious,al2025effect,baek2024effect} of video advertisements.

Among all content types, video advertisements are the most challenging to moderate due to their complexity and the need for precise localization \cite{huang2024lita,sstkd_pami,urur,chen2024timemarker,gu2024context,sstkd} of violations. Recent work, such as RAVEN \cite{raven}, has made significant progress in identifying coarse-grained violation labels and sub-scenes by leveraging large-scale coarsely annotated datasets and fine-tuning with smaller, precisely annotated datasets. However, in practical industrial applications, 
RAVEN’s capabilities still face certain limitations that highlight opportunities for further enhancement: 1) \textbf{Fine-Grained Understanding}: While RAVEN performs well in detecting major violation categories, 
its ability to identify finer-grained sub-categories and localize sub-scenes with high precision remains an area for improvement; 
2) \textbf{Explainability}: RAVEN’s reasoning process occasionally produces explanations that are verbose or misaligned with label rules, and in some cases, its chain-of-thought (CoT) reasoning may contradict the final output, reducing its interpretability; 3) \textbf{Generalization}: RAVEN’s performance in out-of-domain (OOD) scenarios is limited, which restricts its applicability in diverse industrial settings. 

These limitations arise from three key factors: 1) RAVEN’s reliance on fixed rule-based rewards in its GRPO-based \cite{guo2025deepseek} passive reinforcement learning framework, which limits its adaptability to samples of varying difficulty; 2) insufficient constraints on the model’s reasoning and attribution processes, which can lead to inconsistent explanations; 3) a lack of generalized understanding of Ad knowledge and moderation rules, which can result in overfitting to specific labels and reduced OOD performance.

To the end, we propose RAVEN++ with three key innovations: 1) \textbf{Active Reinforcement Learning}: RAVEN++ dynamically rolls out samples of varying difficulty during training and adaptively interleaves supervised fine-tuning (SFT) with reinforcement learning (RL) to optimize performance, leveraging the complementary strengths of SFT for tasks beyond the model’s current capabilities and RL for refining existing skills; 2) \textbf{Fine-Grained Violation Understanding}: By designing a hierarchical set of several reward functions that incorporate Tversky Distance and distill reasoning ability from larger models, RAVEN++ significantly improves the accuracy of fine-grained violation categories, the precision of sub-scene localization, and the reliability and logical coherence of its explanations; 3) \textbf{Progressive Multi-Stage Training}: RAVEN++ introduces a systematic training framework that combines progressive knowledge injection of Ad and moderation rules, curriculum-based passive RL, and active RL, effectively leveraging both large-scale noisy data and small-scale precisely annotated data to enhance violation granularity, generalization, and interpretability. 

We conduct extensive experiments to validate RAVEN++ using both publicly available datasets and proprietary industrial data, on both offline and online deployment scenarios. Our results demonstrate that the RAVEN++ model outperforms models of the same scale, including general-purpose LLMs like LLaVa \cite{llava} and Qwen\cite{qwen}, as well as the specialized video moderation model RAVEN\cite{raven}, in terms of fine-grained violation understanding, reasoning capabilities, and generalization ability.

\begin{figure*}[!ht]
      \centering
      \includegraphics[width=0.8\linewidth]{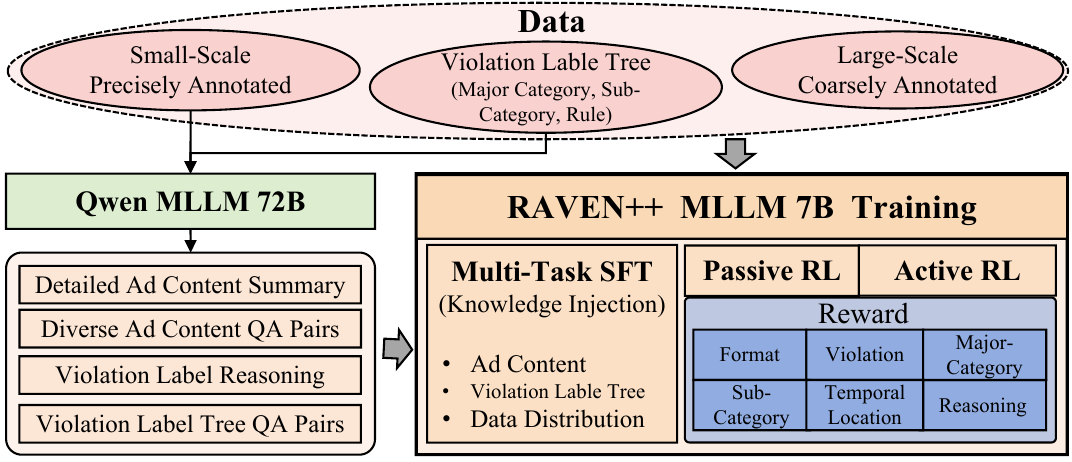}
      \caption{The Overview of RAVEN++ Training.}
      \label{fig_train_overview}
    \end{figure*}

\section{Related Work}

\subsection{Advancements in Reinforcement Learning for Multimodal Systems}

Recent breakthroughs in Multimodal Large Language Models (MLLMs) have revolutionized how machines interpret \cite{llava,qwen_vl,toxvidlm,qwen} and generate content across visual and textual domains \cite{yin2023survey,wang2025multi,ji2024pptformer,llava_cot,xu2024large,zhucpcf,dlpl}. While foundational architectures like CLIP \cite{clip} and BLIP \cite{blip} established strong cross-modal alignment, and subsequent models like Flamingo \cite{flamingo} incorporated temporal awareness, these systems predominantly rely on supervised learning paradigms. This dependence often leads to catastrophic forgetting when adapting to new tasks and limited generalization to unseen data distributions. Emerging research has begun exploring reinforcement learning \cite{zhu2025popen,yu2024rlhf,amini2024direct,song2024preference,liustatistical} with curriculum learning \cite{curriculum_rl,curriculum,graves2017automated,hacohen2019power,soviany2022curriculum,pentina2015curriculum} as a complementary framework to address these limitations. Rather than treating reinforcement learning as a separate optimization mechanism, contemporary approaches integrate it as a core component of the multimodal reasoning pipeline. Techniques such as reward-weighted policy optimization \cite{rlhf} and preference-based learning \cite{dpo} have demonstrated remarkable success in aligning model outputs with human preferences, particularly in domains requiring nuanced temporal understanding. The integration of chain-of-thought reasoning \cite{cot,ji2024tree} with reinforcement learning has further enabled models to decompose complex multimodal tasks into structured decision-making processes, creating more interpretable and controllable systems.

\subsection{Active Learning Paradigms for Content Understanding}

The paradigm of active reinforcement learning represents a significant shift from traditional passive learning approaches. While conventional systems process whatever data they are given, active reinforcement learning empowers models to strategically acquire the most valuable perceptual information through dynamic interaction with their environment. This approach fundamentally transforms the model from a passive observer to an active participant in the learning process. Modern implementations of active reinforcement learning build upon early theoretical foundations \cite{epshteyn2008active} but incorporate sophisticated perceptual capabilities. Contemporary systems \cite{shang2023active} demonstrate that active viewpoint selection and strategic information gathering can dramatically improve sample efficiency and generalization performance. In the context of content moderation \cite{kolla2024llm,kumar2024watch,blackwell2025content,al2025effect}, this active approach enables models to proactively identify subtle violations that might be missed through passive observation alone. The framework allows models to dynamically adjust their perceptual focus, prioritizing regions of interest and temporal segments that are most likely to contain policy violations, thereby optimizing both computational resources and detection accuracy.

\section{Method}

\subsection{Overview}

\subsubsection{Problem Formulation}
Given an input video \( V \), a predefined hierarchical violation labels tree \( T \), and a prompt \( P \), the RAVEN++ framework outputs:  
1) the violation labels \( \mathcal{C} = \{c_1, c_2, \dots, c_n\} \) associated with \( V \),  
2) the sub-categories \( \mathcal{S}_c = \{s_1, s_2, \dots, s_m\} \) for each violation label \( c \),  
3) the temporal intervals \( \mathcal{X}_{c,s} = (t_{c,s}^l, t_{c,s}^r) \) of the sub-scenes corresponding to each sub-category \( s \) of violation \( c \), and  
4) the reasoning \( \mathcal{R}_{c,s} \) behind each sub-category \( s \), providing interpretable explanations for the detected violations.  Here, \( t_{c,s}^l \) and \( t_{c,s}^r \) denote the start and end times of the sub-scene for sub-category \( s \) of violation \( c \), respectively. The hierarchical structure of \( T \) is defined as \textit{“major category \(\rightarrow\) sub-category \(\rightarrow\) rule”}.  This extends RAVEN's \cite{raven} capabilities by incorporating fine-grained reasoning and improved generalization. Specifically, RAVEN++ outputs fine-grained sub-labels for violations and their corresponding reasoning, enabling more detailed and actionable insights.

\subsubsection{Training}
As shown in Fig. \ref{fig_train_overview}, the manually annotated results \( \mathcal{Y}_{c,s} = (y_{c,s}^l, y_{c,s}^r) \) often deviate from the ground truth \( \mathcal{Z}_{c,s} = (z_{c,s}^l, z_{c,s}^r) \) due to annotation errors or ambiguities. While RAVEN achieved significant improvements by leveraging GRPO-based passive reinforcement learning (RL) to handle noisy annotations, we aim to further enhance the framework for: 
1) fine-grained identification of sub-labels \( \mathcal{S}_c \) for each violation \( c \),  
2) precise localization of the temporal intervals \( \mathcal{X}_{c,s} = (t_{c,s}^l, t_{c,s}^r) \) of sub-scenes corresponding to each sub-category \( s \), and  
3) interpretable and reliable reasoning \( \mathcal{R}_{c,s} \) for each violation.

To achieve these goals, RAVEN++ introduces progressive multi-stage training: 1) Knowledge Injection with Augmented Data: the base model is initialized with Ad content and label rule knowledge with augmented data by lightweight SFT;  2) Passive RL with Large-Scale Noisy Data: the model is trained on large-scale coarsely annotated data and a small set of precisely annotated data, guided by a hierarchical set of 6 reward functions to learn global data distributions; 3) Active RL with Small-Scale Precise Data: the model actively identifies valuable samples through designed sampling strategies, enabling comprehensive improvements with minimal precisely annotated data.

\subsection{Stage 1: Knowledge Injection with Augmented Data}

The knowledge injection process integrates both Ad content and moderation rule knowledge into the base model through a unified SFT framework. This process consists of the following steps:  

1) Video Summarization and Ad Knowledge Extraction:  The input video \( V \) is processed by a large model (e.g., Qwen-72B) to generate a detailed caption or summary, including the product name, product attributes (e.g., medicine, cosmetics, clothing), target audience, and key messaging. Based on the summary, Qwen-72B generates question-answer (QA) pairs, such as: 
\vspace{-12pt}
\begin{center}
   \colorbox{mygray}{\scalebox{0.8}{\fbox{
   \begin{minipage}{0.58\textwidth}
   \textit{Question: What product is advertised in this video? \\  
   Answer: The product is a sunscreen named XXX. \\  
   Question: Who is the target audience for this advertisement? \\  
   Answer: The target audience is young adults aged about 18–60.}
   \end{minipage}
   }}}
   \end{center}  

2) Hierarchical Rule Knowledge Extraction:  
   The moderation rules are structured hierarchically. QA pairs are generated to capture the relationships between major categories, sub-categories, and rules, such as:  
    \vspace{-12pt}
   \begin{center}
   \colorbox{mygray}{\scalebox{0.8}{\fbox{
   \begin{minipage}{0.58\textwidth}
   \textit{Question: What are the sub-categories and rules for the main category ‘Discomforting Content’? \\  
   Answer: Sub-labels: ‘Gory Content,’ ‘...’, Rules: ... \\  
   Question: What constitutes a violation under the sub-category ‘Misleading Claims’? \\  
   Answer: Claims that exaggerate product efficacy without evidence.}
   \end{minipage}
   }}}
   \end{center}  

3) Joint SFT:  
   All the generated QA pairs are combined with a small portion of precisely annotated data to fine-tune the base model.

\subsection{Reward Design for RL}

The reward function $R$ optimizes the model’s RL performance across six dimensions,

\subsubsection{Format Reward \( R_{\text{format}} \)}

   It ensures the model’s output adheres to the predefined structure, 
   which is critical for downstream processing and interpretation,
   The format requires the output to include the following components: 
   \begin{equation}
   \small
   \begin{aligned} 
        & R_{\text{format}}  = \mathbb{I}( 
        \langle  \text{think} \rangle \langle\text{/think} \rangle \langle  \text{reason} \rangle \text{content summarization:   } \\
        & \text{... risk analysis: ... conclusion: ...}   \langle\text{/reason} \rangle \langle \text{violation}\rangle \text{Y/N}  \\ 
      & \langle\text{/violation}\rangle \langle  \text{result}  \rangle   \text{\{major: ..., sub: ..., ground: ...\}}\langle\text{/result}\rangle \\ 
      &  \langle \text{result}  \rangle  \text{\{major: ..., sub: ..., ground: ...\}}\langle\text{/result}\rangle ...),
   \end{aligned}
   \label{eq_format}
   \end{equation}
   where \( \mathbb{I}(\cdot) \) is an indicator function.

   \subsubsection{Violation Reward \( R_{\text{violation}} \)}
   
   This reward evaluates whether the model’s prediction of a violation matches the ground truth (GT):

   \begin{equation}
   \begin{aligned}
   R_{\text{violation}} = \mathbb{I}(P_{\text{violation}} = GT_{\text{violation}}).~~
   \end{aligned}
   \end{equation}

  \subsubsection{Reasoning Consistency Reward \( R_{\text{reason}} \)}

   This reward encourages logical and coherent reasoning chains \( \mathcal{R}_{c,s} \), structured into summarization, risk analysis, and conclusion. It compares the model’s CoT reasoning with a high-quality structured CoT from Qwen-72B (Eq. \ref{eq_format}), using a semantic cosine similarity measure with BERT \cite{bert} to enhance interpretability and reliability.
   \begin{equation}
   \begin{aligned}
   R_{\text{reason}} = \text{Sim}(\text{CoT}_{\text{model}}, \text{CoT}_{\text{Qwen-72B}}).~~
   \end{aligned}
   \end{equation}

\subsubsection{Major Category Reward \( R_{\text{major}} \)}

   This reward measures the accuracy of the model’s prediction for the major violation category \( c \in \mathcal{C} \). It uses the Tversky distance \cite{tversky}, a generalization of the Dice coefficient, to evaluate the similarity between the predicted and ground truth major categories:  
   \begin{equation}
   \begin{aligned}
   R_{\text{major}} = 1 - \text{Tversky}(P_{\text{major}}, GT_{\text{major}}), ~~
   \end{aligned}
   \end{equation}  
   the Tversky distance is particularly effective for imbalanced datasets, as it allows for flexible weighting of false positives and false negatives.  

 \subsubsection{Sub-Category Reward \( R_{\text{sub}} \)}
 
   Similar to the major category reward, this component evaluates the model’s performance in identifying the specific violation sub-category \( s \in \mathcal{S}_c \). It also employs the Tversky distance for robust evaluation:  
   \begin{equation}
   \begin{aligned}
   R_{\text{sub}} = 1 - \text{Tversky}(P_{\text{sub}}, GT_{\text{sub}}).
   \end{aligned}
   \end{equation}

\subsubsection{Temporal Grounding Reward \( R_{\text{ground}} \)}

   It evaluates the model’s ability to localize the violation temporal sub-scenes \( \mathcal{X}_{c,s} = (t_{c,s}^l, t_{c,s}^r) \) within the video. Following RAVEN, it combines the Intersection over Union (IoU) metric, which measures the overlap between predicted and ground truth temporal segments, with a boundary alignment reward that penalizes deviations in the start and end times to ensure precise localization: 
   \begin{equation}
       \begin{aligned}
           R_{\text{ground}} =  & \text{IoU}(P_{\text{ground}}, GT_{\text{ground}}) + \\
    & \text{Boundary}(P_{\text{ground}}, GT_{\text{ground}}).
       \end{aligned}
   \end{equation}

The overall reward \( R \) is 
\begin{equation}
\begin{aligned}
R = \lambda_1 R_{\text{format}} + \lambda_2 R_{\text{violation}} + \lambda_3 R_{\text{major}} + \\
 \lambda_4 R_{\text{sub}} + \lambda_5 R_{\text{ground}} + \lambda_6 R_{\text{reason}},
\end{aligned}
\end{equation} 
\( \lambda_1, \lambda_2, \lambda_3, \lambda_4, \lambda_5, \lambda_6 \) are dynamically adjustable hyperparameters that control the relative importance of each reward component.  

\begin{figure}[!ht]
      \centering
      \includegraphics[width=1\linewidth]{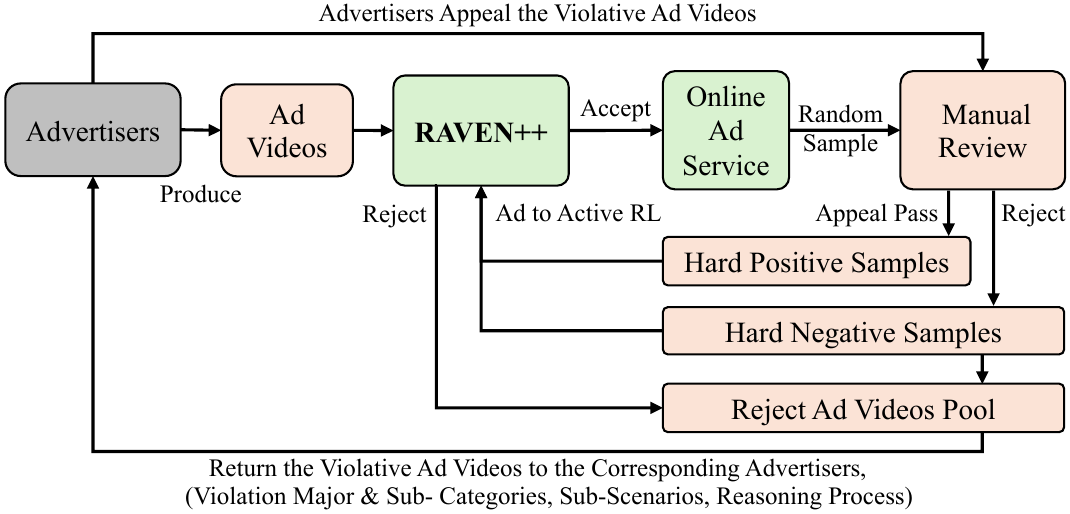}
      \caption{The deployment of RAVEN++.}
      \label{fig_deployment}
    \end{figure}

\subsection{Stage 2: Curriculum-Based Passive RL with Large-Scale Noisy Data}
To effectively train the model, we adopt a curriculum-based passive reinforcement learning strategy, which dynamically adjusts the weights of the reward components across three phases. This approach ensures the model progressively learns from simpler to more complex tasks, aligning with its evolving capabilities.  

\subsubsection{Phase 1: Format and Violation Learning}   

In the early phase, the model focuses on mastering the output format and accurately identifying whether a violation exists. This is critical because the task format is complex, and errors in format or violation detection can propagate to downstream tasks. To prioritize these aspects, the weights \( \lambda_1 \) (for \( R_{\text{format}} \)) and \( \lambda_2 \) (for \( R_{\text{violation}} \)) are set to relatively large values, while the weights for other reward components are kept small (the reward weights are [1, 1, 0.5, 0.3, 0, 0.1]). This stage ensures the model builds a strong foundation in adhering to the required structure and correctly classifying violations.  

\subsubsection{Phase 2: Fine-Grained Violation Detection and Reasoning}

Once the model demonstrates proficiency in format adherence and violation detection, the focus shifts to more challenging tasks, including identifying major categories \( \mathcal{C} \), sub-categories \( \mathcal{S}_c \), and generating interpretable reasoning \( \mathcal{R}_{c,s} \). In this stage, the weights \( \lambda_1 \) and \( \lambda_2 \) are gradually reduced, while the weights \( \lambda_3 \) (for \( R_{\text{major}} \)), \( \lambda_4 \) (for \( R_{\text{sub}} \)), and \( \lambda_6 \) (for \( R_{\text{reason}} \)) are increased (the reward weights are [0.5, 0.5, 1, 1, 0, 0.5]). This transition enables the model to refine its ability to perform fine-grained violation detection and provide coherent explanations.

\subsubsection{Phase 3: Temporal Grounding} 

In the final phase, the model tackles the most challenging task: precise temporal localization of violations \( \mathcal{X}_{c,s} = (t_{c,s}^l, t_{c,s}^r) \). At this stage, the weight \( \lambda_5 \) (for \( R_{\text{ground}} \)) is significantly increased, while the weights for other components are maintained or slightly reduced (the reward weights are [0.2, 0.2, 1, 1, 1, 0.5]). This ensures the model dedicates sufficient resources to learning the temporal boundaries of violations, which is critical for accurate sub-scene localization.

\begin{table*}[]
\centering
\scalebox{0.68}{\begin{tabular}{c|cc|cc|cc|cc|cc|cc}
\toprule
 \multirow{2}{*}{\textbf{Method}}                                                 & \multicolumn{2}{c|}{\begin{tabular}[c]{@{}c@{}}\textbf{Marketing} \\ \textbf{Exaggerate} \end{tabular}} & \multicolumn{2}{c|}{\begin{tabular}[c]{@{}c@{}}\textbf{Discomforting}\\\textbf{Content} \end{tabular}} & \multicolumn{2}{c|}{\begin{tabular}[c]{@{}c@{}}\textbf{Vulgar}\\ \textbf{Content} \end{tabular}} & \multicolumn{2}{c|}{\begin{tabular}[c]{@{}c@{}}\textbf{Requiring}\\ \textbf{Credential Review} \end{tabular}} & \multicolumn{2}{c|}{\begin{tabular}[c]{@{}c@{}}\textbf{Prohibited}\\ \textbf{Goods/Services} \end{tabular}}            & \multicolumn{2}{c}{\begin{tabular}[c]{@{}c@{}}\textbf{Average}\\ \textbf{Major Category} \end{tabular}} \\ \cmidrule{2-13} 
                          & {\color[HTML]{333333} Cate.(P/R)}        & {\color[HTML]{333333} Gro.} & {\color[HTML]{333333} Cate.(P/R)}     & {\color[HTML]{333333} Gro.}     & {\color[HTML]{333333} Cate.(P/R)}  & {\color[HTML]{333333} Gro.} & {\color[HTML]{333333} Cate.(P/R)}        & {\color[HTML]{333333} Gro.}        & {\color[HTML]{333333} Cate.(P/R)} & {\color[HTML]{333333} Gro.} & Cate.(P/R)        & Gro.       \\ \midrule

{\begin{tabular}[c]{@{}c@{}}Small\\ Models \end{tabular} } & 0.681/0.532  & - & 0.707/0.679  &  -  &  0.667/0.654  &  -   & 0.711/0.687 &  -  & 0.721/0.734    & - &  0.697/0.657     &  -       \\ \midrule
{\begin{tabular}[c]{@{}c@{}}LLaVA\\ -v1.5-SFT \end{tabular} } & 0.796/0.756  & 0.398 & 0.798/0.772  &  0.385  &  0.771/0.799  &  0.400    & 0.754/0.701 &  0.432  & 0.789/0.761    & 0.567 &  0.782/0.758     &  0.436       \\ \midrule
{\begin{tabular}[c]{@{}c@{}}Qwen2.5-VL\\ -7B-SFT \end{tabular} } & 0.832/0.787  & 0.424 & 0.821/0.798  &  0.402  &  0.800/0.810  &  0.411    & 0.773/0.702 &  0.461  & 0.797/0.771    & 0.580 &  0.805/0.774    &  0.456       \\ \midrule
{RAVEN} & 0.851/0.801  & 0.521 & 0.843/0.812  &  0.477  &  0.810/0.831  &  0.565    & 0.802/0.713 &  0.541  & 0.825/0.784    & 0.669 &  0.826/0.788     &  0.555 \\ \midrule
{\textbf{RAVEN++}} & 0.905/0.840  & 0.601 & 0.889/0.859  &  0.562  &  0.862/0.870  &  0.650    & 0.873/0.738 &  0.671  & 0.859/0.832    & 0.741 &  \textbf{0.878/0.828}     &  \textbf{0.647} 
\\ \bottomrule
\end{tabular}}
\caption{Performance of Violation Major Category (Precision/Recall) and Violation Temporal Grounding (mIoU) on Industrial  Dataset. ``Cate." indicates ``Category'', and ``Gro." indicates ``Grounding''.}
\label{table_sota_industry}
\end{table*}

\begin{table*}[h]
\centering
\scalebox{0.75}{\begin{tabular}{c|c|c|c|c|c|c}
\toprule
\textbf{Method} & \begin{tabular}{@{}c@{}}\textbf{Marketing}\\\textbf{Exaggerate}\end{tabular} & \begin{tabular}{@{}c@{}}\textbf{Discomforting}\\\textbf{Content}\end{tabular} & \begin{tabular}{@{}c@{}}\textbf{Vulgar}\\\textbf{Content}\end{tabular} & \begin{tabular}{@{}c@{}}\textbf{Requiring}\\\textbf{Credential Review}\end{tabular} & \begin{tabular}{@{}c@{}}\textbf{Prohibited}\\\textbf{Goods/Services}\end{tabular} & \begin{tabular}{@{}c@{}}\textbf{Average} \\ \textbf{Sub-Category (P/R)}\end{tabular} \\
\midrule
RAVEN & 0.638/0.585 & 0.601/0.590 & 0.588/0.611 & 0.586/0.567 & 0.596/0.579 & 0.602/0.586 \\\midrule
RAVEN++ & 0.735/0.654 & 0.705/0.622 & 0.668/0.652 & 0.659/0.607 & 0.682/0.704 & \textbf{0.690/0.650} \\
\bottomrule
\end{tabular}}
\caption{Performance of Violation Sub-Category (Precision/Recall) on Industrial Dataset.}
\label{table_sota_industry2}
\end{table*}

\subsection{Stage 3: Active RL with Small-Scale Precise Data}

In the final training phase of RAVEN++, we introduce an active reinforcement learning (RL) stage, which leverages a small amount of high-quality, finely annotated data to perform targeted replay training on valuable samples. This stage is designed to maximize data efficiency while significantly improving the model’s performance. During training, we interleave supervised fine-tuning (SFT) and RL, maintaining two separate data buffers for each approach. This strategy is motivated by our observation that RL excels at maintaining and improving performance on tasks within the model’s current capabilities, while SFT is more effective at enabling progress on tasks beyond the model’s current scope, which is also articled in \cite{relift,chusft}.

\textbf{Operation:} To operationalize this, we dynamically sample and group training examples based on their complexity and the model’s performance, assigning them to either the SFT or RL buffer as follows:  
\begin{itemize}
    \item \textbf{Fundamental Knowledge Gaps}: Samples where the model fails to identify violations or misclassifies the major category are assigned to the SFT buffer. These cases represent fundamental knowledge gaps that require targeted supplementation to address the model’s inability to recognize violations or major categories. 
    \item \textbf{Refinement of Existing Capabilities}: Samples where the model correctly identifies the major category but misclassifies the sub-category, exhibits low IoU for sub-scene localization, or deviates significantly from Qwen-72B’s reasoning are assigned to the RL buffer with larger weights. These cases represent tasks within the model’s current scope but require refinement in sub-category classification, temporal grounding, or reasoning consistency. 
    \item \textbf{Standard Cases}: Samples that do not fall into the above categories are processed using normal RL execution but are assigned a very small weight in the reward function. This ensures that the model maintains its performance on tasks it already handles well, without overfitting to less challenging examples.  
This dynamic sampling strategy ensures that SFT is used for knowledge supplementation on tasks that exceed the model’s current capabilities, while RL is applied to refine performance on tasks within the model’s current scope.  
\end{itemize}

\textbf{Training Details and Data Efficiency:}   
The active RL stage places a high emphasis on data quality, as it relies on a small number of finely annotated samples. To avoid overfitting and ensure stable learning, we use a small learning rate during this phase. When the buffer accumulates enough challenging questions to form a training batch, we perform an SFT or RL step using these examples. This ensures that the model is trained on a diverse and representative set of challenging cases, enabling it to address its weaknesses effectively. Remarkably, even with this limited data, we observe rapid performance improvements, demonstrating the data efficiency of this approach. This efficiency is attributed to the targeted replay training, which focuses on the most valuable and challenging samples, enabling the model to quickly address its weaknesses and enhance its strengths.

\section{Deployment}

As shown in Fig. \ref{fig_deployment}, based on the deployment of RAVEN, RAVEN++ integrates difficult positive and negative samples identified by online reviewers into the Active RL stage, enabling incremental learning and continuous model improvement. This enhancement allows the system to adapt dynamically to emerging violation patterns and improve its precision over time.

\begin{table}[]
\centering
\scalebox{0.8}{\begin{tabular}{c|cc}
\toprule
\multirow{2}{*}{\textbf{Method}}                                      & \multicolumn{2}{c}{\textbf{Average}}           \\ \cmidrule{2-3} 
                                                             & Cate. (P/R)          & Gro.           \\ \midrule
LLaVA-v1.5-SFT   & 0.509/0.501          & 0.370          \\ \midrule
Qwen2.5-VL-7B-SFT & 0.537/0.517          & 0.384          \\ \midrule
RAVEN                                               & 0.551/0.530 & 0.435 \\ \midrule
\textbf{RAVEN++}                                               & \textbf{0.584/0.551} & \textbf{0.487}  \\ \bottomrule
\end{tabular}}
\caption{Performance of Violation Category (Precision/Recall) and Violation Temporal Grounding (mIoU) on Public MultiHateClip Dataset.}
\label{study_hate}
\end{table}

\section{Experiments and Results}

We conduct extensive experiments from both offline testing and online testing, utilizing both public dataset and practical industrial dataset. 
For fair comparison, we follow the dataset settings in RAVEN.

\subsection{Datasets} \label{sec_dataset}

Our experimental setup follows the dataset settings in RAVEN \cite{raven}. The industrial dataset consists of 38,000 training videos (with mixed precise and coarse annotations) and 5,000 precisely annotated test videos. All annotations adhere to the same six major violation categories and temporal interval labels defined in RAVEN. For public testing, we use the available subset of the MultiHateClip \cite{wang2024multihateclip} dataset, consistent with the RAVEN benchmark.

\subsection{Offline Testing}

We evaluate RAVEN++ against several baselines, including LLaVA-v1.5 \cite{llava}, Qwen2-VL-7B \cite{qwen_vl}, Qwen2.5-VL-7B \cite{qwen_vl}, and RAVEN. The results in Table \ref{table_sota_industry}, \ref{table_sota_industry2} and \ref{study_hate} show that RAVEN++ outperforms all baselines in category accuracy and grounding precision, achieving significant improvements over RAVEN in sub-scene interval localization and reasoning consistency. These gains highlight the effectiveness of RAVEN++'s active RL stage. Furthermore, we also perform independent-samples t-tests, which confirm that the performance differences between RAVEN and RAVEN++ are statistically significant (p < 0.02).

\begin{table}[]
\centering
\scalebox{0.8}{\begin{tabular}{c|cc}
\toprule
\multirow{2}{*}{\textbf{Model}} & \multicolumn{2}{c}{\textbf{Online Sample Average}} \\ \cmidrule{2-3} 
                       & Cate.(P/R)               & Gro.                 \\ \midrule
Small Models          & 0.711/0.668         & -         \\ \midrule
Qwen2.5-VL-7B-SFT             & 0.800/0.787         & 0.478         \\ \midrule
RAVEN       & 0.821/0.803         &   0.563       \\ \midrule
RAVEN++     & \textbf{0.873/0.847}         &   \textbf{0.662}       \\ 
\bottomrule
\end{tabular}}
\caption{A/B Test on the Online Serving.}
\label{exp_online}
\end{table}

\subsection{Online A/B Testing}

To evaluate real-world applicability, we perform a day-long online A/B test on a live business platform, adhering to the framework established by RAVEN. By dedicating 20\% of the total traffic for assessment, we compare RAVEN++ with RAVEN, a compact legacy model, and Qwen2.5-VL-7B-SFT. As shown in Table \ref{exp_online}, RAVEN++ delivers substantial gains in identifying violative videos, achieving higher precision and recall in category detection compared to the legacy model. Moreover, RAVEN++ achieves an 9.9\% improvement over RAVEN in temporal interval localization accuracy.

\begin{table}[]
\centering
\scalebox{0.8}{\begin{tabular}{c|c|c}
\toprule
\textbf{Method} & \begin{tabular}[c]{@{}c@{}}In-Domain\\ (Average Gro.) \end{tabular}  & \begin{tabular}[c]{@{}c@{}}Out-of-Domain\\ (Average Gro.) \end{tabular} \\ \midrule
Qwen2.5-VL-7B-SFT   &   0.433   &   0.246         \\ \midrule
RAVEN  &    0.546      &  0.408          \\ \midrule
RAVEN++  &  0.631      &  0.463          \\ 
\bottomrule
\end{tabular}}
\caption{Study on Generalization Capabilities.}
\label{study_general}
\end{table}

\subsection{Study on Generalization Capabilities}

To further evaluate the generalization ability of RAVEN++, we follow the experimental design of RAVEN, training the model on three in-domain categories (Discomforting Content, Marketing Exaggeration, Requiring Credential Review) and testing it on two out-of-domain categories (Vulgar Content, Prohibited Goods/Services). The results, presented in Table \ref{study_general}, reveal that RAVEN++ consistently achieves higher accuracy and better generalization compared to both the Qwen2.5-VL SFT model and RAVEN. 

\begin{table}[]
\centering
\scalebox{0.85}{\begin{tabular}{c|c|c|c}
\toprule
Multi-Task SFT & Passive RL & Active RL & Gro.  \\ \midrule
$\checkmark$              &            &           & 0.461 \\ \midrule
$\checkmark$              & $\checkmark$          &           & 0.587 \\ \midrule
$\checkmark$              & $\checkmark$          & $\checkmark$         & 0.647 \\ \bottomrule
\end{tabular}}
\caption{Study on Progressive Training Stages.}
\label{study_train}
\end{table}

\begin{table}[]
\centering
\scalebox{0.8}{\begin{tabular}{c|c|c}
\toprule
\begin{tabular}[c]{@{}c@{}}Tversky Distance \\ Reward\end{tabular} & \begin{tabular}[c]{@{}c@{}}Reasoning \\ Reward\end{tabular} & \begin{tabular}[c]{@{}c@{}}Average Major\\ Category (P/R)\end{tabular} \\ \midrule
$\checkmark$                                                                  &                                                             & 0.869/0.820                                                            \\ \midrule
                                                                   & $\checkmark$                                                           & 0.860/0.812                                                            \\ \midrule
$\checkmark$                                                                  & $\checkmark$                                                           & 0.878/0.828                                                            \\ \bottomrule
\end{tabular}}
\caption{Study on  Reward Functions.}
\label{study_reward}
\end{table}

\subsection{Study on  Progressive Training Stages}

Table \ref{study_train} presents ablation studies on the three-stage training process used in this work. The results show that RAVEN++ achieves steady and significant performance improvements with each progressive training stage. Notably, the final active RL stage boosts grounding performance by 6\% compared to the previous stage.  

\subsection{Study on Reward Functions}

RAVEN++ introduces two novel reward functions, Tversky Distance based and Reasoning Reward, which are evaluated through ablation studies on the Industrial dataset. As shown in Table \ref{study_reward}, both functions significantly enhance performance, validating their effectiveness compared to RAVEN.

\section{Conclusion}

We address the critical challenges of video Ad moderation by introducing RAVEN++, a novel framework that significantly enhances fine-grained violation understanding, reasoning capabilities, and generalization ability. Through the integration of Active RL, Fine-Grained Violation Understanding, and Progressive Multi-Stage Training, RAVEN++ achieves superior performance on both offline and online deployment scenarios.

\section{Limitations}

A key limitation of this work is its exclusive focus on the video modality for Ad content moderation. Real-world advertising is inherently multi-modal, combining text, images, and video. Our current approach, which analyzes video in isolation, cannot capture cross-modal inconsistencies—such as misleading text overlaid on benign visuals. This restricts its applicability to video-dominant scenarios. However, this constraint outlines a clear path for future work. Our method provides a foundation for extension towards a unified multi-modal framework. Future efforts will focus on integrating text and image analysis to build a holistic system for robust, comprehensive ad moderation.

\section{Ethical Statement}

Our research adheres to ethical principles and prioritizes user rights. The dataset samples are for scientific analysis only and do not reflect the authors' views. All resources are intended for scientific research purposes only, contributing to the development of more secure and reliable digital platforms.

\bibliography{custom}

\end{document}